\documentclass[letterpaper]{article} 
\usepackage{aaai2026}  
\usepackage{times}  
\usepackage{helvet}  
\usepackage{courier}  
\usepackage[hyphens]{url}  
\usepackage{graphicx} 
\usepackage{natbib}  
\usepackage{caption} 
\usepackage{algorithm}
\usepackage{algorithmic}
\usepackage{tabularx}
\usepackage{multirow}
\usepackage{subcaption}
\usepackage{amsfonts}
\usepackage{amssymb}
\usepackage{amsmath}
\usepackage{booktabs}
\usepackage{xcolor}
\usepackage{multicol}
\usepackage{float}

\usepackage{multibib}
\newcites{SM}{References}

\usepackage{newfloat}
\usepackage{listings}
\DeclareCaptionStyle{ruled}{labelfont=normalfont,labelsep=colon,strut=off} 
\floatstyle{ruled}
\newfloat{listing}{tb}{lst}{}
\floatname{listing}{Listing}
\title{DBGroup: Dual-Branch Point Grouping for Weakly Supervised \\ 3D Semantic Instance Segmentation}
\author{
    Xuexun Liu\textsuperscript{\rm 1}\thanks{Email:2019043026@email.szu.edu.cn},
    Xiaoxu Xu\textsuperscript{\rm 1}, 
    Qiudan Zhang\textsuperscript{\rm 1}\thanks{Corresponding author.},
    Lin Ma\textsuperscript{\rm 2},
    Xu Wang\textsuperscript{\rm 1}
}
\affiliations{
    \textsuperscript{\rm 1}College of Computer Science and Software Engineering, Shenzhen University, Shenzhen, 518060, China.\\

    \textsuperscript{\rm 2}Meituan Inc., China. \\
}

\usepackage{bibentry}

\begin{document}

\maketitle

\begin{abstract}
Weakly supervised 3D instance segmentation is essential for 3D scene understanding, especially as the growing scale of data and high annotation costs associated with fully supervised approaches. Existing methods primarily rely on two forms of weak supervision: one-thing-one-click annotations and bounding box annotations, both of which aim to reduce labeling efforts. However, these approaches still encounter limitations, including  labor-intensive annotation processes, high complexity, and reliance on expert annotators. To address these challenges, we propose \textbf{DBGroup}, a two-stage weakly supervised 3D instance segmentation framework that leverages scene-level annotations as a more efficient and scalable alternative. In the first stage, we introduce a Dual-Branch Point Grouping module to generate pseudo labels guided by semantic and mask cues extracted from multi-view images. To further improve label quality, we develop two refinement strategies: Granularity-Aware Instance Merging and Semantic Selection and Propagation. The second stage involves multi-round self-training on an end-to-end instance segmentation network using the refined pseudo-labels. Additionally, we introduce an Instance Mask Filter strategy to address inconsistencies within the pseudo labels. Extensive experiments demonstrate that DBGroup achieves competitive performance compared to sparse-point-level supervised 3D instance segmentation methods, while surpassing state-of-the-art scene-level supervised 3D semantic segmentation approaches. Code is available at https://github.com/liuxuexun/DBGroup.
\end{abstract}

\vspace{-0.5cm}

\section{Introduction}

3D instance segmentation is a fundamental task in 3D scene understanding that aims to predict both masks and semantic categories for individual objects within point cloud scenes. This field has garnered significant attention in recent years due to its extensive applications in the emerging domains such as embodied AI, robotics, and AR/VR technologies \cite{less, xu2025weakly, li2022expansion}. Recently, learning-based 3D instance segmentation methods~\cite{mask3d, pbnet} have achieved remarkable performance. However, these methods heavily rely on time-consuming and labor-intensive point-wise annotations for both semantic and instance labels.

\begin{figure}[t]
\centering
\includegraphics[width=\linewidth]{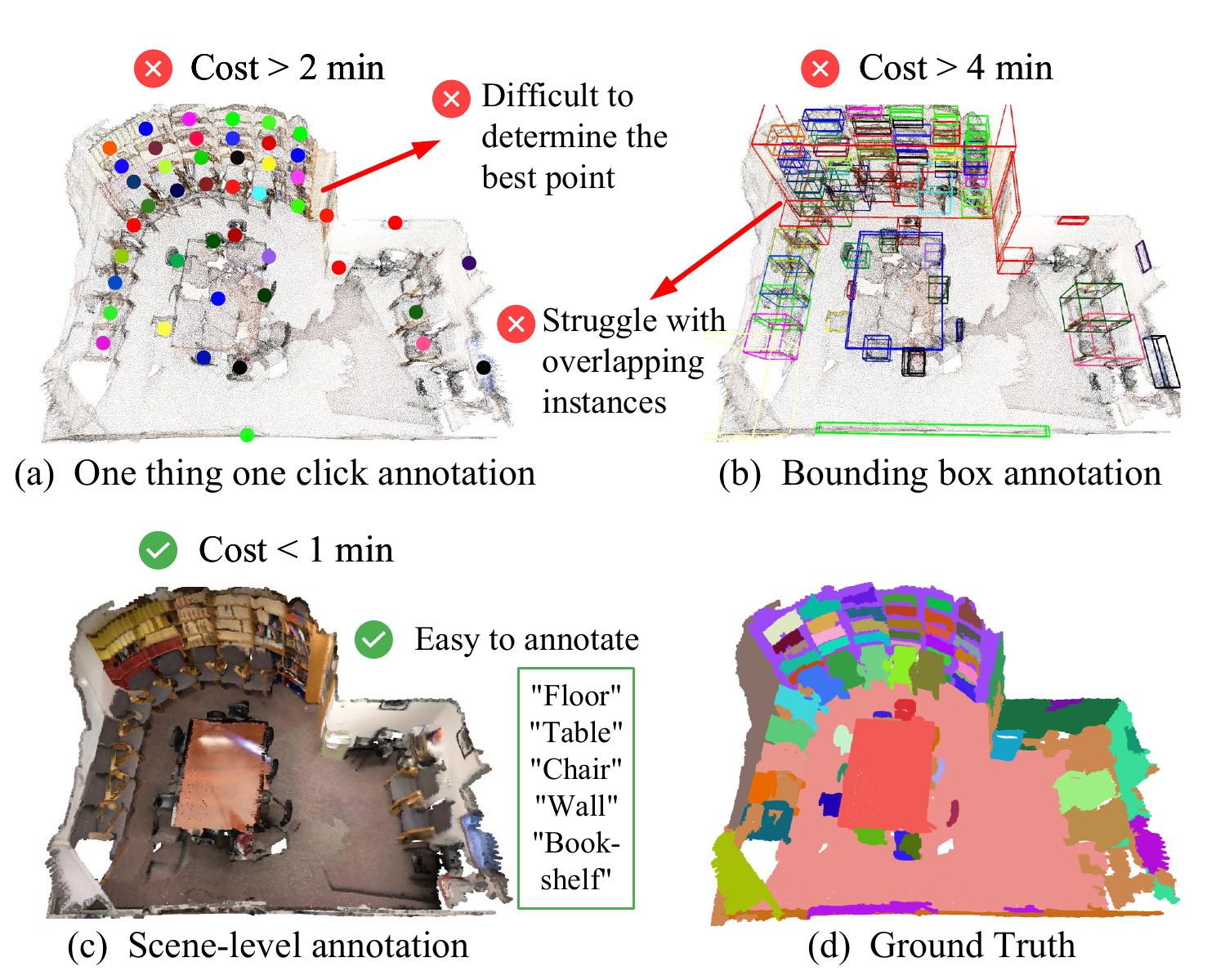}
   \caption{Comparison between conventional weak annotation formats and our proposed scene-level annotation in 3D instance segmentation task. \vspace{-0.5cm}}
\label{fig:introduction}
\end{figure}

To alleviate this limitation, numerous methods~\cite{seggroup, WSIS,box2mask, bsnet} have explored weakly supervised 3D instance segmentation. As shown in Fig.~\ref{fig:introduction}, these approaches primarily utilize two types of weak annotations: ``one thing one click" (OTOC) annotation and bounding box (BBox) annotation . OTOC indicates each instance will be annotated at least one point while BBox need to annotate each instance with bounding box. While these annotation methods substantially reduce annotation costs, there still exist some issues. First, they still require distinguishing individual instances, which fails to significantly reduce the time-consuming annotation process. Second, they often demand specialized annotator training.  For OTOC, annotators must precisely determine click point positions for complex structures; for BBox, untrained annotators struggle with overlapping instances with precise boundaries.

Therefore, inspired by recent weakly supervised 3D semantic segmentation approaches~\cite{mprm, 3DSSVLG}, we propose to employ scene-level annotation for weakly supervised 3D instance segmentation. As illustrated in Fig.~\ref{fig:introduction} (c), scene-level annotation merely requires labeling the categories of objects present within a scene. Compared to OTOC and BBox annotation, scene-level annotation offers several significant advantages. On the one hand, it substantially reduces annotation time, requiring less than 1 minute per scene on average, whereas OTOC and BBox annotations typically require more than 2 minutes and 4 minutes~\cite{mil}. On the other hand, it eliminates the need for professional training of annotators, as they only need to distinguish the categories of objects in the scene. 

\begin{figure}
\centering
\includegraphics[width=\linewidth]{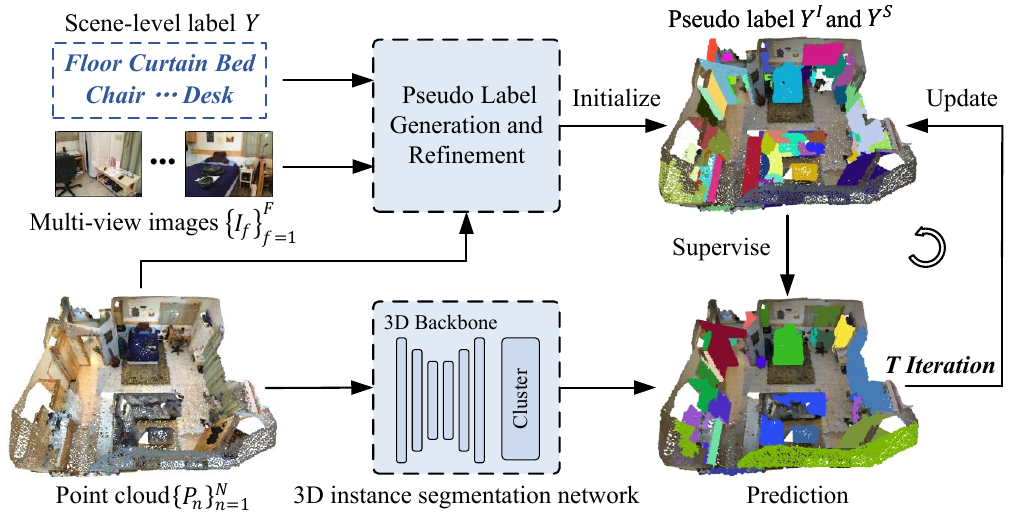}
   \caption{Overview of our weakly supervised 3D segmentation pipeline. It includes a Pseudo Label Generation and Refinement module, as well as a multi-round self-training 3D instance segmentation network. \vspace{-0.5cm}}
\label{fig:overview}
\end{figure}

\begin{figure*}
\centering
\includegraphics[width=\linewidth]{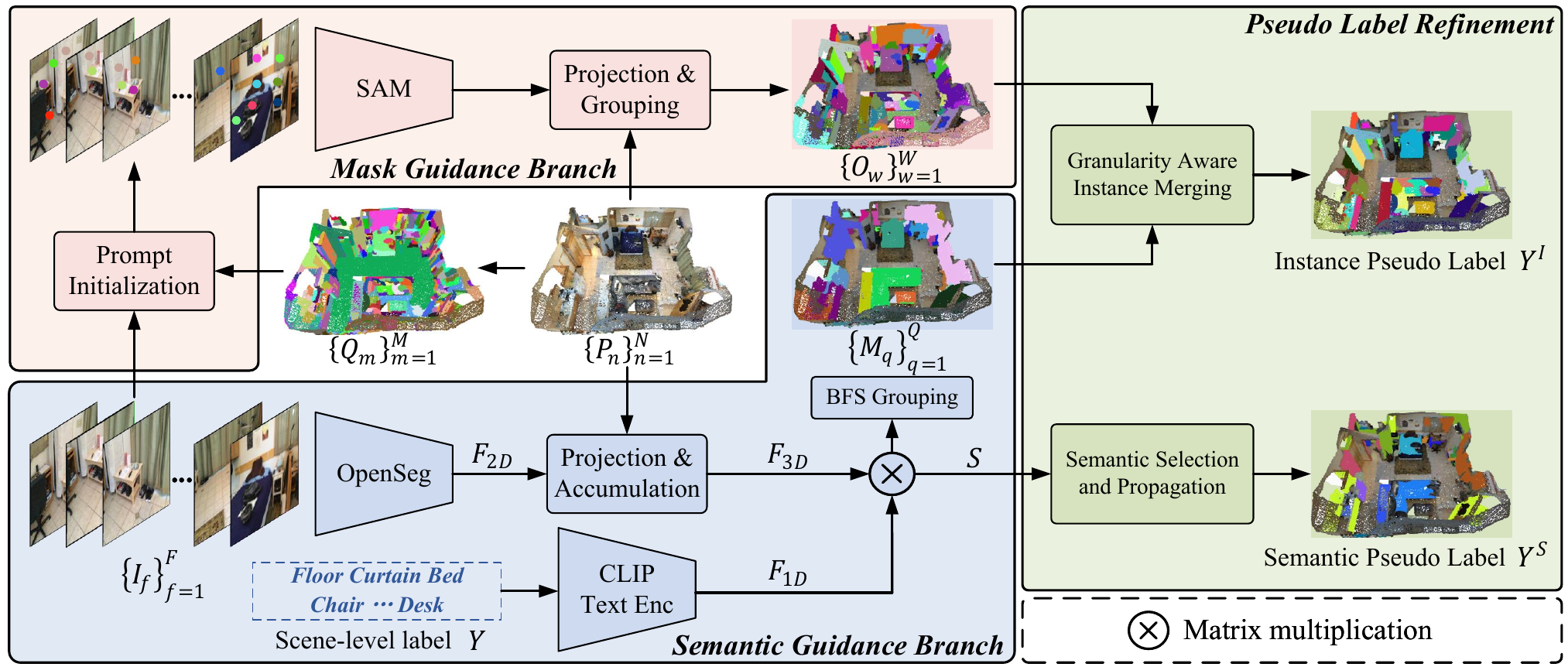}
   \caption{The workflow of our designed Pseudo Label Generation and Refinement. We propose a dual-branch point grouping architecture and two pseudo label refinement strategies. The SGB extracts features from multi-view images via a pre-trained 2D model, projects them into 3D point clouds, computes similarities using text encoder features, and applies BFS clustering to generate coarse-grained masks. The MGB projects superpoints onto multi-view images as SAM prompts, creating masks that group 3D points into fine-grained masks. GAIM merges or splits masks for instance pseudo labels, while SSP filters semantic scores for semantic pseudo labels. }
\label{fig:framework}
\end{figure*}

However, compared with OTOC and BBox annotations, scene-level annotation lacks instance-level information, making it difficult to guide the model in segmenting individual instances. To address this issue, as illustrated in Fig.~\ref{fig:overview}, we propose a two-stage paradigm comprising Pseudo Label Generation and Refinement, followed by a multi-round self-training 3D instance segmentation network. In the first stage (see Fig.~\ref{fig:framework}), we introduce a Dual-Branch Point Grouping module consisting of a Semantic Guidance Branch (SGB) and a Mask Guidance Branch (MGB). The SGB leverages a pre-trained vision-language model to extract pseudo semantic labels, and applies a radius-based Breadth-First Search (BFS) clustering algorithm to generate coarse-grained pseudo instance labels. Concurrently, the MGB employs superpoint prompts and the Segment Anything Model (SAM)~\cite{sam} to produce 2D masks, which are then used to guide the grouping of 3D point clouds into fine-grained clusters. To further enhance pseudo label quality, we introduce two strategies: Granularity-Aware Instance Merging (GAIM) and Semantic Selection and Propagation (SSP). 
The GAIM strategy performs instance splitting or merging based on mask intersection ratios, while the SSP strategy filters out low-confidence predictions and propagates high-confidence predictions to broader regions. Furthermore, to mitigate inconsistencies between separately processed instance and semantic pseudo labels, we incorporate an Instance Mask Filter strategy in the second stage (Fig.~\ref{fig:network}), enabling the model to better focus on individual instances.

To summarize, our contributions are as follows:

\begin{itemize}
    \item We propose a new weakly supervised 3D semantic instance segmentation method DBGroup, which use only scene-level labels. To the best of our knowledge, DBGroup is the first work investigating 3D instance segmentation with scene-level labels.
    \item  Our DBGroup leverages a Dual-Branch Point Grouping module to produce both pseudo instance and semantic label and utilizes the Pseudo Label Refinement to generate high-quality pseudo labels. Moreover, an Instance Mask Filter strategy is introduced to improve the consistency of instance predictions.
    \item Extensive experiments demonstrate that our approach achieves competitive performance compared to sparse-point-level supervised 3D instance segmentation methods, while utilizing more label-efficient scene-level annotations. Moreover, our method surpasses current state-of-the-art scene-level supervised 3D semantic segmentation approaches. These results validate the effectiveness and superiority of our proposed framework.
\end{itemize}

\section{Related work}

{\textbf{\textit{Weakly Supervised 3D Segmentation.}}} In response to the challenge of manual annotation, weakly supervised 3D segmentation has garnered significant research interest in recent years. Researchers typically leverage multiple weakly supervised signals. Box-level annotation offers the most comprehensive representation of instance and geometric details, the primary challenge of these methods~\cite{box2mask, WISGP, gapro, bsnet} is accurately assigning points within overlapping regions of 3D bounding boxes. In  sparse-points-level annotation, some methods~\cite{hybridcr, pointmatch, hierarchical} adopt the mean teacher paradigm, while others~\cite{seggroup, learning, OTOC, coll} employ graph-based approaches to propagate sparse supervised signals. Methods~\cite{mprm, WYPR, mil} based on scene-level annotation predominantly employ 3D Class Activation Mapping (CAM)~\cite{cam} for semantic object localization. Subsequent approaches~\cite{joint, MIT} enhance this framework by incorporating 2D features for cross-modal alignment and fusion. Recent state-of-the-art advance~\cite{3DSSVLG, xuclassaware} leverage the robust generalization capabilities of 2D visual-language models to align textual information of scene label with 3D visual features. Although scene-level annotation is cost-effective and easily implementable, current instance segmentation approaches are unable to effectively utilize such coarse-grained labeling schemes.

{\textbf{ \textit{2D Pre-trained Model in 3D Segmentation.}}} With the rapid advancement of 2D pre-trained models, researchers have increasingly explored their application to 3D segmentation tasks. Some approaches~\cite{sam3d, sai3d, sampro3d} leverage SAM~\cite{sam} for class-agnostic segmentation of 3D point clouds, while others~\cite{openscene, li2025cross, openmask3d} utilize CLIP~\cite{clip} to align point cloud features with text embeddings for open-vocabulary segmentation. Recent developments have extended these 2D pre-trained frameworks to weakly supervised 3D segmentation. 3DSSVLG~\cite{3DSSVLG} pioneered the use of CLIP models to guide weakly supervised semantic segmentation through textual semantics, while REAL~\cite{oracle} employs SAM as an active learning annotator to iteratively refine segmentation masks. However, most of these methods require 2D images as input during training or inference, which inadvertently increases computational demands and limits potential applications. In contrast, our two-stage method only requires processing the dataset containing 2D images once.

\section{Methodology}

\textbf{\textit{Problem Definition.}} In our weakly supervised setting, as shown in Fig.~\ref{fig:framework}, the inputs comprise three components: \textit{\textbf{1)}} a point cloud scene $\{\mathbf{P}_{n}\}_{n=1}^{N} \in \mathbb{R}^{N\times6}$ containing $N$ points with its spatial coordinates and color values attributes $(x,y,z,r,g,b)$, \textit{\textbf{2)}} its corresponding RGB multi-view images $\{\mathbf{I}_{f}\}_{f=1}^{F}$ containing $F$ frames with a resolution of $H \times W$, \textit{\textbf{3)}} scene-level label $Y\in\{Floor, Curtain, Bed,...,Desk\}^{K}$, $K$ denotes the number of categories in the scene. Our objective is to produce high quality pseudo instance and semantic label $Y^{I} \in \mathbb{R}^{N}$ and $Y^{S} \in \mathbb{R}^{N}$ to facilitate the $T$-iteration self-training.


\subsection{Semantic Guidance Branch}
\label{section:Semantic Guidance Branch}

Inspired by previous works~\cite{openscene, 3DSSVLG}, we first leverage pre-trained vision-language models~\cite{clip, openseg} to obtain multi-modal embeddings. By performing matrix multiplication between these embeddings, we then derive pseudo semantic labels for the point cloud. Finally, we propose a BFS Grouping strategy to aggregate points that belong to the same instance and generate the predicted coarse-grained instance masks.

\textbf{\textit{Feature Extraction.}} Following~\cite{openscene, 3DSSVLG}, we utilize the pre-trained image encoder and the text encoder from OpenSeg \cite{openseg} to obtain the embeddings $\mathbf{F_{2D}} \in \mathbb{R}^{F\times H\times W\times C}$ and $\mathbf{F_{1D}} \in \mathbb{R}^{K\times C}$, respectively, where $C$ denotes the embedding dimension.

{\textbf{\textit{Projection and Accumulation.}}} For each point $\mathbf{P}_{n}$ in the point cloud $\{\mathbf{P}_{n}\}_{n=1}^{N}$, we first project it onto all multi-view images using geometric camera calibration matrices. After applying visibility filtering, we obtain the corresponding 2D positions of $\mathbf{P}_{n}$ on the relevant images. These positions are then used to extract the corresponding embeddings from $\mathbf{F_{2D}}$. Since a single point $\mathbf{P}_{n}$ may be visible in multiple images, we aggregate its multiple associated embeddings by computing their mean. This embedding extraction and aggregation process is repeated for all points in the point cloud, resulting in a aggregated 3D embeddings $\mathbf{F_{3D}} \in \mathbb{R}^{N\times C}$.

{\textbf{\textit{BFS Grouping.}}} Given the scene-level label embeddings $\mathbf{F_{1D}} \in \mathbb{R}^{K\times C}$ and the aggregated 3D embeddings $\mathbf{F_{3D}} \in \mathbb{R}^{N\times C}$, we first compute classification scores $\mathbf{S} \in \mathbb{R}^{N \times K}$ through matrix multiplication. Based on $\mathbf{S}$, we adopt a widely-used Breadth-First Search (BFS) algorithm \cite{pointgroup} to group points belonging to foreground categories. Specifically, we select an initial foreground anchor point $\mathbf{P}_{n}$ and search for neighboring points within a spherical region with radius $r$ centered at $\mathbf{P}_{n}$. Neighboring points that sharing the same semantic prediction as the anchor point are grouped together into the same cluster. Subsequently, the newly grouped points serve as anchors to iteratively repeat the above procedure until no additional points satisfying the criteria can be included. Upon completion of this iterative process, we obtain a complete instance cluster, which is then removed from the original point cloud. We subsequently select another anchor point to perform clustering for the next instance, repeating this procedure until all foreground points have been clustered into their respective instances.

After preceding steps, our SGB produces final classification scores $\mathbf{S}$ and coarse-grained instance masks $\{\mathbf{M}_{q}\}_{q=1}^{Q}$, where each mask is a binary mask $\mathbf{M}_{q} \in \{0,1\}^{N}$.

\subsection{\textbf{Mask Guidance Branch}}
\label{section:Mask Guidance Branch}

While the coarse-grained instance masks $\{\mathbf{M}_{q}\}_{q=1}^{Q}$ from SGB capture complete object instances, they often lack sufficient precision. This is primarily due to the radius-based clustering performed by the BFS algorithm, which may mistakenly merge two distinct instances of the same category into a single cluster when their spatial distance is smaller than the predefined radius. To overcome this issue, we introduce a Mask Guidance Branch that leverages superpoint prompts and SAM~\cite{sam} to produce fine-grained instance masks $\{\mathbf{O}_{w}\}_{w=1}^{W}$ for enhancing the coarse-grained masks.

{\textbf{\textit{Prompt Initialization.}}} Given a point cloud scene $\{\mathbf{P}_{n}\}_{n=1}^{N}$, we first oversegment the points into superpoints $\{\mathbf{Q}_m\}_{m=1}^M$ using a normal-based graph cut algorithm \cite{oversegmentaion}, where $M$ denotes the number of superpoints in $\{\mathbf{P}_{n}\}_{n=1}^{N}$. For each superpoint $\mathbf{Q}_m$, we compute its centroid $\mathbf{c}_m \in \mathbb{R}^3$ as the point closest to the average coordinates of all points within $\mathbf{Q}_m$. We project the resulting centroid set $\{\mathbf{c}_m\}_{m=1}^M$ onto the multi-view images, serving as prompts for SAM to perform preliminary mask segmentation. Compared to point-level prompt initialization~\cite{sampro3d}, which treats every point in the point cloud as an individual prompt, this superpoint-based strategy significantly reduces computational overhead. Moreover, using a small number of geometrically meaningful prompts enables SAM to generate more accurate and fine-grained instance masks. Finally, we can obtain 2D segmentation masks for all frames $\{\mathbf{MS}_{f}\}_{f=1}^{F}$.

{\textbf{\textit{Projection and Grouping.}}} 
Given the superpoint centroid set $\{\mathbf{c}_m\}_{m=1}^M$ with previous $\{\mathbf{MS}_{f}\}_{f=1}^{F}$, we conduct a multi-view projection to achieve comprehensive 3D scene segmentation. Each 3D point is projected onto all masked frames $\{\mathbf{MS}_{f}\}_{f=1}^{F}$, with labels assigned through a robust voting mechanism. Specifically, if a point $\mathbf{P}_{n}$ falls within the mask of prompt $\mathbf{c}_m$ in a given frame $\mathbf{MS}_{f}$, $\mathbf{P}_{n}$ is assigned the instance ID $m$ at frame $f$. The final instance ID of $\mathbf{P}_{n}$ is determined as the most frequently assigned prompt ID across all $F$ frames. Through this process, we obtain fine-grained instance masks $\{\mathbf{O}_{w}\}_{w=1}^{W}$, where each mask is represented as a binary vector $\mathbf{O}_{w} \in \{0,1\}^{N}$.

\subsection{\textbf{Granularity Aware Instance Merging}}
\label{section:GAIM}

{\bf \textit{Granularity Aware Assignment.}} Through the SGB and MGB, we obtain the coarse-grained instance masks $\{\mathbf{M}_{q}\}_{q=1}^{Q}$ and fine-grained instance masks $\{\mathbf{O}_{w}\}_{w=1}^{W}$. The coarse-grained masks are better suited for capturing complete instance objects, whereas the fine-grained masks are more effective at accurately delineating object boundaries. To integrate these complementary representations, we propose a Granularity Aware Assignment strategy. We first construct an overlap matrix $\mathbf{A} \in \mathbb{Z}^{Q \times W}$ to quantify the intersections between the two sets of masks. For each coarse-grained mask $\mathbf{M}_{q}$, we identify the fine-grained mask $\mathbf{O}_{w}$ with the highest overlap ratio $\rho$. If $\rho$ exceeds a predefined threshold $\theta$, the coarse-grained mask is retained. Otherwise, we consider the coarse-grained mask $\mathbf{M}_{q}$ to be under-segmented and likely to contain multiple distinct instances. To address this, we compute the intersection between the coarse-grained mask and all fine-grained masks $\mathbf{O}_{w}$, and retain only the overlapping regions with fine-grained masks to refine and decompose the under-segmented coarse-grained region. After that, we can obtain the ensemble instance masks $\Gamma$. The pseudo code is shown in our supplementary material.

{\bf \textit{Small Instances Merging.}} Although Granularity Aware Assignment can effectively decompose under segmented masks, fine-grained masks also have the problem of over segmentation. To resolve this challenge, we leverage spatial adjacency relationships between instances to facilitate appropriate merging. Our method first evaluates inter-instance adjacency relationship by employing the K-Nearest Neighbors (KNN) algorithm for all points in the scene. The adjacency relationship between instances A and B is established based on the proximity of their points. Specifically, the nearest neighboring instance of instance A is defined as the instance containing the greatest number of points that are nearest neighbors to points within instance A. Subsequently, we calculate the point number within each instance and merges instances containing fewer points than a predefined threshold $\gamma$ with their nearest large neighboring instance. This process ultimately yields the instance pseudo label $\mathbf{Y^{I}} \in \mathbb{R}^{N}$. Given that instance centroids are subject to spatial offset, with smaller instances particularly vulnerable to such displacement. We avoid using centroids for determining inter-instance adjacency relationship.

\subsection{\textbf{Semantic Selection and Propagation}}
\label{section:SSP}

{\bf \textit{Semantic Selection.}} 
As is well known, low-confidence pseudo labels are more likely to be inaccurate compared to high-confidence ones, which also holds true for the classification score matrix $\mathbf{S}$ generated by the SGB. To address this issue, we propose a semantic selection algorithm. Specifically, for each category present in a scene, we retain only the top $\alpha\%$ of points with the highest classification confidence and discard predictions for the remaining points in that category. Unlike conventional global top-$\alpha\%$ filtering based solely on classification scores, our class-wise selection strategy maintains semantic balance between majority and minority classes, effectively alleviating performance degradation caused by the under-representation of minority classes.

{\bf \textit{Superpoint Propagation.}} After the selection process, the semantic pseudo labels become sparse with limited coverage rate and lack spatial continuity. Therefore, we propose a Superpoint Propagation algorithm that leverages geometric priors for effective label propagation. Specifically, we utilize superpoints $\{\mathbf{Q}_{m}\}_{m=1}^{M}$ to partition the point set $\{\mathbf{P}_{n}\}_{n=1}^{N}$ into $M$ distinct groups. Within each group, we determine the most frequently predicted category and propagate this label to all points within the corresponding superpoint yielding the semantic pseudo label $\mathbf{Y^{S}} \in \mathbb{R}^{N}$.

\subsection{Self-Training}
\label{section:self}

To gradually improve the quality of pseudo label and enhance the performance on the validation set, we adopt a $T$-iteration self-training strategy, as illustrated in Fig.~\ref{fig:overview}. This iterative approach allows the model to generate increasingly accurate and robust predictions by using its progressively refined outputs as supervisory signals.

\begin{figure}
\centering
\includegraphics[width=\linewidth]{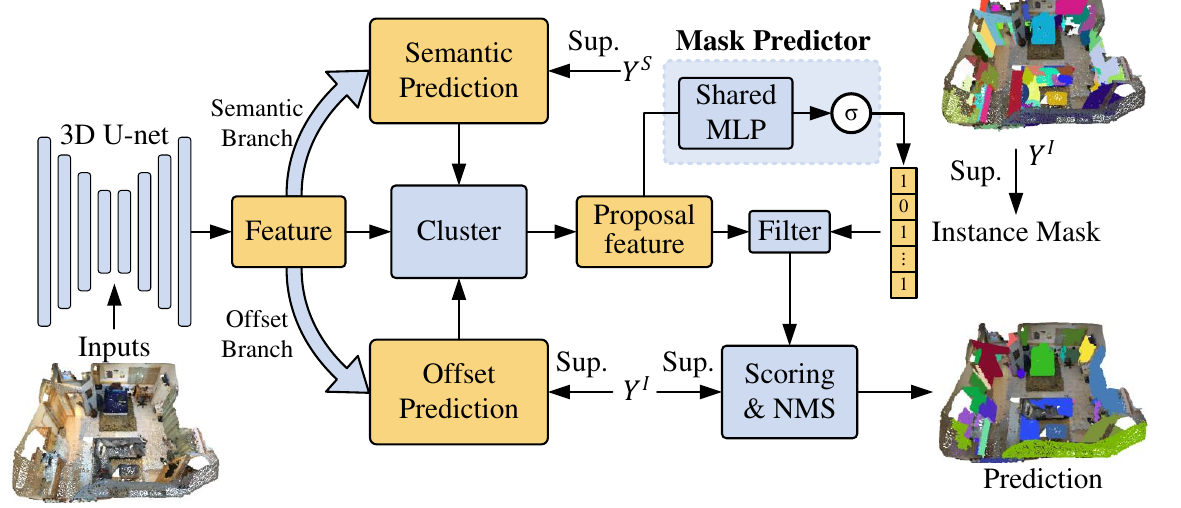}
   \caption{Framework of our 3D instance segmentation network. Following a grouping-based paradigm, we use a 3D U-Net for feature extraction with semantic and offset branches to predict semantic labels and offsets. After clustering, an Instance Mask Filter removes irrelevant points from proposals, with final predictions obtained via a scoring network and non-maximum suppression (NMS). \vspace{-0.5cm}}
\label{fig:network}
\end{figure}


The network architecture is shown in Fig.~\ref{fig:network}. We adopt a grouping-based 3D instance segmentation framework following~\cite{pointgroup, softgroup, pbnet}. More details are shown in our supplementary material.

However, due to the independence of the SGB and MGB, the generated semantic pseudo labels $\mathbf{Y^{S}}$ and instance pseudo labels $\mathbf{Y^{I}}$ may be inconsistent, introducing noise into the clustering process. To address this issue, we propose an \textit{Instance Mask Filter} strategy. Specifically, for each proposal feature $PF_i \in \mathbb{R}^{N_i \times C}$, where $N_i$ denotes the number of points in the $i$-th proposal, we apply a shared multi-layer perceptron (MLP) followed by binary thresholding at 0.5 to predict a proposal-specific mask $PM_i \in \{0,1\}^{N_i}$. This mask is used to filter the proposal embeddings $PF_i$, refining the instance representation. Following~\cite{pbnet, dknet}, we supervise $PM_i$ using a combination of binary cross-entropy loss $L_{\text{bce}}$ and Dice loss~\cite{dice} $L_{\text{dice}}$.

\section{Experiment}
{\bf \textit{Datasets.}} We evaluate our method on two widely-used indoor point cloud datasets: ScanNetV2~\cite{scannet} and S3DIS~\cite{s3dis}. ScanNetV2 comprises 1,513 training samples across 20 semantic categories, with instance segmentation required for 18 categories. Following the default setting, we use 1,201 scenes for training and 312 scenes for validation. S3DIS consists of 271 scenes distributed across 6 areas, encompassing 13 semantic categories that all require instance segmentation. We follow the common evaluation setting, using Area 5 as the validation set and the remaining areas for training. Both datasets provide RGB-D sequences along with intrinsic and extrinsic camera parameters, enabling 3D-to-2D projection. 

{\bf \textit{Evaluation Metrics.}} We evaluate our approach on both semantic segmentation and instance segmentation tasks. For semantic segmentation, we employ mean Intersection over Union (mIoU) as the evaluation metric. For instance segmentation, we utilize mean Average Precision (mAP) at various Intersection over Union (IoU) thresholds: 0.25 (AP$_{25}$), 0.5 (AP$_{50}$), and the average over the range [0.5:0.95:0.05] (AP). Additionally, for the S3DIS dataset, we report mean Recall (mRec) and mean Precision (mPre) at an IoU threshold of 0.5. 

{\bf \textit{Implementations Details.}} For the hyper-parameters of our pseudo label refinement module, we set the BFS Grouping radius $r$ to 0.04, overlap threshold $\theta$ to 0.4, the point threshold $\gamma$ to 200 and the top-$\alpha\%$ threshold to 30\% respectively. In our training process, we set the number of self-training iterations $T$ to 3,  employed the Adam optimizer~\cite{adam} with a batch size of 6 and initialized the learning rate at 0.001. We implement a cosine annealing schedule~\cite{sgdr}, beginning with 50 epochs and applying per-step decay, for a total of 256 training epochs. Our architecture utilized MinkowskiNet34C~\cite{mink} as the 3D network backbone, and all experimental evaluations were conducted using three NVIDIA RTX 4090 GPUs. 


\begin{table}[]
\centering
\caption{Comparison of different methods for 3D instance segmentation on ScanNetV2 dataset. ``mask" represents fully annotation. ``box" indicates bounding box annotation. ``$x\%$ point" denotes the percentage of annotated points relative to all points in the entire dataset. ``scene" denotes scene-level annotation. }
\begin{tabular}{ccccc}
\toprule
{\color[HTML]{000000} Method}     & {\color[HTML]{000000} Annotation}  & {\color[HTML]{000000} AP}   & {\color[HTML]{000000} $\text{AP}_{50}$} & {\color[HTML]{000000} $\text{AP}_{25}$} \\ \midrule
{\color[HTML]{000000} GSPN}       & {\color[HTML]{000000} mask}         & {\color[HTML]{000000} 19.3} & {\color[HTML]{000000} 37.8} & 53.4                        \\
{\color[HTML]{000000} 3D-SIS}     & {\color[HTML]{000000} mask}         & {\color[HTML]{000000} -}    & {\color[HTML]{000000} 18.7} & {\color[HTML]{000000} 35.7} \\
{\color[HTML]{000000} MTML}       & {\color[HTML]{000000} mask}         & {\color[HTML]{000000} 20.3} & {\color[HTML]{000000} 40.2} & {\color[HTML]{000000} 55.4} \\
{\color[HTML]{000000} PointGroup} & {\color[HTML]{000000} mask}         & {\color[HTML]{000000} 34.8} & {\color[HTML]{000000} 56.9} & {\color[HTML]{000000} 71.3} \\
{\color[HTML]{000000} SoftGroup}  & {\color[HTML]{000000} mask}         & {\color[HTML]{000000} 46.0} & {\color[HTML]{000000} 67.6} & {\color[HTML]{000000} 78.9} \\
{\color[HTML]{000000} Mask3D}     & {\color[HTML]{000000} mask}         & {\color[HTML]{000000} 55.2} & {\color[HTML]{000000} 73.7} & {\color[HTML]{000000} -}    \\ \midrule
{\color[HTML]{000000} SPIB}       & {\color[HTML]{000000} box}          & {\color[HTML]{000000} -}    & {\color[HTML]{000000} 38.6} & {\color[HTML]{000000} 61.4} \\
{\color[HTML]{000000} Box2Mask}   & {\color[HTML]{000000} box}          & {\color[HTML]{000000} 39.1} & {\color[HTML]{000000} 59.7} & {\color[HTML]{000000} 71.8} \\
{\color[HTML]{000000} BSNet}      & {\color[HTML]{000000} box}          & {\color[HTML]{000000} 53.3} & {\color[HTML]{000000} 72.7} & {\color[HTML]{000000} 83.4} \\ \midrule
{\color[HTML]{000000} CSC-50}        & {\color[HTML]{000000} 0.034\% point} & {\color[HTML]{000000} 22.9}    & {\color[HTML]{000000} 41.4} & {\color[HTML]{000000} 62.0}    \\
{\color[HTML]{000000} TWIST}      & {\color[HTML]{000000} 5\% point}   & {\color[HTML]{000000} 27.0} & {\color[HTML]{000000} 44.1} & {\color[HTML]{000000} 56.2} \\
{\color[HTML]{000000} SegGroup}   & {\color[HTML]{000000} 0.02\% point} & {\color[HTML]{000000} 23.4} & {\color[HTML]{000000} 43.4} & {\color[HTML]{000000} 62.9} \\
{\color[HTML]{000000} 3D-WSIS}    & {\color[HTML]{000000} 0.02\% point} & {\color[HTML]{000000} 28.1} & {\color[HTML]{000000} 47.2} & {\color[HTML]{000000} 67.5} \\ \midrule
{\color[HTML]{000000} DBGroup (Ours)}       & {\color[HTML]{000000} scene}        & {\color[HTML]{000000} 28.6} & {\color[HTML]{000000} 46.2} & {\color[HTML]{000000} 59.6} \\ \bottomrule 
\end{tabular}
\vspace{-0.5cm}
\label{tab:main_scannet_ins}
\end{table}

\begin{table}[]
\centering
\caption{Comparison of different methods for 3D instance segmentation on S3DIS dataset.}
\setlength{\tabcolsep}{0.45mm}{
\begin{tabular}{ccccccc}
\toprule
{\color[HTML]{000000} Method}   & {\color[HTML]{000000} Annotation}  & {\color[HTML]{000000} AP}   & {\color[HTML]{000000} AP$_{50}$}  & mPrec                       & mRec                        \\ \midrule
PointGroup                     & {\color[HTML]{000000} mask}         & {\color[HTML]{000000} -}    & 57.8                            & 61.9                        & 62.1                        \\
{\color[HTML]{000000} Mask3D }   & {\color[HTML]{000000} mask}         & {\color[HTML]{000000} 56.6} & {\color[HTML]{000000} 68.4}     & {\color[HTML]{000000} 68.7} & {\color[HTML]{000000} 66.3} \\ \midrule
{\color[HTML]{000000} Box2Mask }   & {\color[HTML]{000000} box}         & {\color[HTML]{000000} 43.6} & {\color[HTML]{000000} 54.6}     & {\color[HTML]{000000} 66.7} & {\color[HTML]{000000} 65.5} \\
{\color[HTML]{000000} BSNet }   & {\color[HTML]{000000} box}         & {\color[HTML]{000000} 53.0} & {\color[HTML]{000000} 64.3}     & {\color[HTML]{000000} -} & {\color[HTML]{000000} -} \\ \midrule
{\color[HTML]{000000} SegGroup } & {\color[HTML]{000000} 0.02\% point} & 21.0                        & 29.8                                                & {\color[HTML]{000000} 47.2} & {\color[HTML]{000000} 34.9} \\
{\color[HTML]{000000} 3D-WSIS }  & {\color[HTML]{000000} 0.02\% point} & {\color[HTML]{000000} 23.3} & {\color[HTML]{000000} 33.0}  & {\color[HTML]{000000} 50.8} & {\color[HTML]{000000} 38.9} \\ \midrule
{\color[HTML]{000000} DBGroup (Ours)}     & {\color[HTML]{000000} scene}        & {\color[HTML]{000000} 27.1}    & {\color[HTML]{000000} 40.5}    & {\color[HTML]{000000} 38.5}       & {\color[HTML]{000000} 44.8}   \\ \bottomrule
\end{tabular}}
\label{tab:main_s3dis_ins}
\end{table}

\subsection{Comparison with SOTAs}

{\textbf{\textit{Evaluation on 3D Instance Segmentation.}}} As shown in Tab.~\ref{tab:main_scannet_ins} and \ref{tab:main_s3dis_ins}, we evaluate our method against a range of approaches utilizing different levels of supervision. The results demonstrate that our DBGroup achieves competitive performance on both benchmarks, outperforming the majority of point-annotation-based methods. Specifically, when compared to 3D-WSIS~\cite{WSIS}, a state-of-the-art method based on point annotations, DBGroup achieves comparable results on the ScanNetV2 dataset, while showing substantial gains on the S3DIS dataset surpassing 3D-WSIS by 3.8\% and 7.5\% in terms of \textit{AP} and \textit{AP$_{50}$}, respectively. It is worth noting that scene-level annotations, required by DBGroup, are significantly more cost-efficient than point-level annotations, underscoring the effectiveness of our approach under simpler supervision.

Furthermore, DBGroup demonstrates lower performance in the \textit{mPrec} metric compared to point-annotation-based methods, while significantly outperforming them in \textit{mRec}. This trade-off arises from the nature of scene-level supervision: lacking fine-grained spatial cues, pseudo labels generated from scene-level annotations tend to include irrelevant regions, thereby lowering precision. In contrast, point-level annotations offer precise localization but limited coverage, causing the model to overlook unlabeled object parts and thus suffer in recall. Our results highlight DBGroup’s ability to strike a better balance between annotation cost and segmentation performance.

\begin{table}[]
\centering
\caption{Comparison of different methods for 3D semantic segmentation on ScanNetV2 and S3DIS dataset.}
\scalebox{1}{
\begin{tabular}{cccc}
\toprule
{\color[HTML]{000000} }                         & {\color[HTML]{000000} }                              & {\color[HTML]{000000} ScanNetV2} & {\color[HTML]{000000} S3DIS}     \\
\multirow{-2}{*}{{\color[HTML]{000000} Method}} & \multirow{-2}{*}{{\color[HTML]{000000} Annotation}} & {\color[HTML]{000000} mIoU} & {\color[HTML]{000000} mIoU} \\ \midrule
{\color[HTML]{000000} PointNet }                 & {\color[HTML]{000000} mask}                          & {\color[HTML]{000000} -}         & {\color[HTML]{000000} 41.1}      \\
{\color[HTML]{000000} MinkowskiNet }             & {\color[HTML]{000000} mask}                          & {\color[HTML]{000000} 72.2}      & {\color[HTML]{000000} 65.4}      \\
{\color[HTML]{000000} PTv3}       & {\color[HTML]{000000} mask}                          & 78.6                             & 74.3                             \\ \midrule
{\color[HTML]{000000} HybridCR }                 & 1\% point                                            & 56.9                             & 65.3                             \\
{\color[HTML]{000000} OTOC++ }                   & {\color[HTML]{000000} 0.02\% point}                  & {\color[HTML]{000000} 70.5}      & {\color[HTML]{000000} 56.6}      \\ \midrule
{\color[HTML]{000000} MPRM }                     & {\color[HTML]{000000} scene}                         & 24.4                             & {\color[HTML]{000000} 10.3}      \\
{\color[HTML]{000000} MIL-Trans }                & {\color[HTML]{000000} scene}                         & 26.2                             & 12.9                             \\
{\color[HTML]{000000} WYPR }                     & {\color[HTML]{000000} scene}                         & {\color[HTML]{000000} 29.6}      & 22.3                             \\
{\color[HTML]{000000} MIT }                      & {\color[HTML]{000000} scene}                         & {\color[HTML]{000000} 35.8}      & 27.7                             \\
{\color[HTML]{000000} 3DSS-VLG }                  & {\color[HTML]{000000} scene}                         & {\color[HTML]{000000} 49.7}      & {\color[HTML]{000000} 45.3}      \\
{\color[HTML]{000000} DBGroup (Ours)}                     & {\color[HTML]{000000} scene}                         & {\color[HTML]{000000} \textbf{56.9}}      & {\color[HTML]{000000} \textbf{48.2}}         \\ \bottomrule
\end{tabular}}
\vspace{-0.5cm}
\label{tab:main_sem}
\end{table}

{\textbf{\textit{Evaluation on 3D Semantic Segmentation.}}} We further evaluate DBGroup on the 3D semantic segmentation task and compare it with existing methods under different annotation settings. As shown in Tab.~\ref{tab:main_sem}, when trained with only scene-level annotations, DBGroup surpasses the previous state-of-the-art by 7.2\% on ScanNetV2 and 2.9\% on the S3DIS dataset. These results clearly demonstrate the effectiveness of our method for 3D semantic segmentation.

\subsection{Ablation Study}

In this section, we conduct a series of ablation studies to validate the effectiveness of each component in our proposed framework. All experiments are performed on the ScanNetV2 dataset, using the same experimental settings as described in previous sections, except for the specific module under evaluation. To ensure training efficiency and facilitate fair comparisons, all models are trained with a single iteration ($T=1$) unless otherwise stated.

\begin{table}[]
\centering
\caption{Quantitative ablation results of SGB, MGB and GAIM.}
\begin{tabular}{cl|ccc}
\toprule
                           & {\color[HTML]{000000} Setting}        & {\color[HTML]{000000} AP}   & {\color[HTML]{000000} AP$_{50}$} & {\color[HTML]{000000} AP$_{25}$} \\ \midrule
{\color[HTML]{000000} (a)} & {\color[HTML]{000000} Only SGB}       & {\color[HTML]{000000} 16.3} & {\color[HTML]{000000} 30.7} & {\color[HTML]{000000} 53.3} \\
{\color[HTML]{000000} (b)} & {\color[HTML]{000000} Only MGB}       & {\color[HTML]{000000} 15.5} & {\color[HTML]{000000} 30.2} & {\color[HTML]{000000} 56.2} \\
{\color[HTML]{000000} (c)} & {\color[HTML]{000000} MGB+SGB (GAIM)} & {\color[HTML]{000000} \textbf{17.4}} & {\color[HTML]{000000} \textbf{32.7}} & {\color[HTML]{000000} \textbf{58.7}} \\
{\color[HTML]{000000} (d)} & {\color[HTML]{000000} MGB+SGB (BM)}   & {\color[HTML]{000000} 16.3} & {\color[HTML]{000000} 30.8} & {\color[HTML]{000000} 54.0} \\ \bottomrule
\end{tabular}
\vspace{-0.5cm}
\label{tab:ablation_GAIM}
\end{table}

\begin{figure*}
\centering
\includegraphics[width=\linewidth]{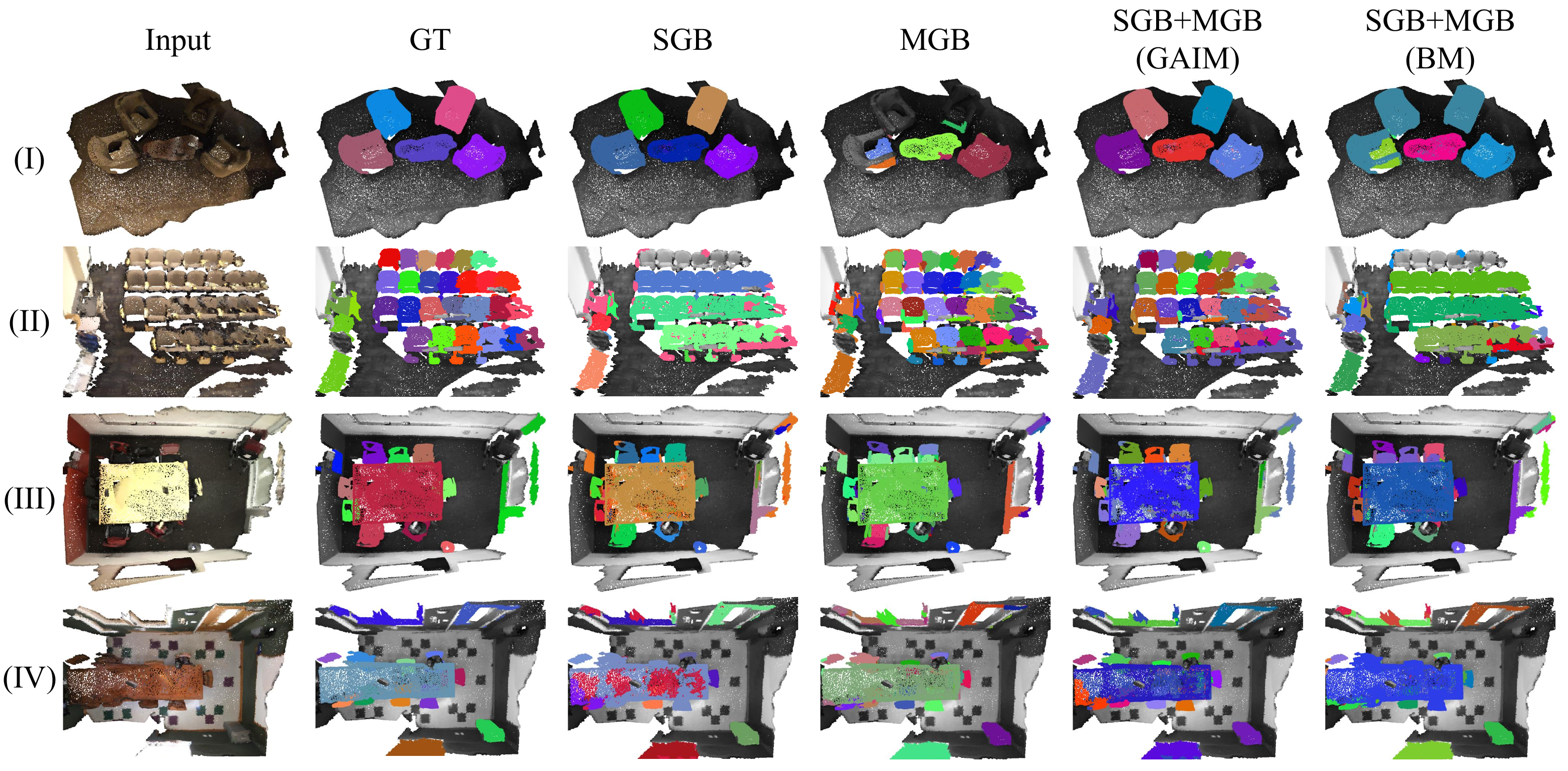}
   \caption{Qualitative ablation results of SGB, MGB and GAIM. From left to right: input point clouds, ground truth, setting (a) to (d) in Tab.\ref{tab:ablation_GAIM}.}
\label{fig:ablation}
\end{figure*}

{\bf \textit{Effect of SGB, MGB, and GAIM.}} In Tab.~\ref{tab:ablation_GAIM}, we present a comparative analysis of how different branches and merging strategies affect the quality of pseudo instance labels. Settings (a) and (b) independently employ the SGB and MGB to generate pseudo instance labels. Setting (c) utilizes both branches to produce instance masks of varying granularities, which are then integrated using our proposed GAIM strategy. Setting (d) implements the Bidirectional Merging (BM) approach from SAM3D~\cite{sam3d} for label integration, which is a widely adopted technique for combining diverse mask predictions into coherent results. Among all approaches examined, our GAIM strategy demonstrates superior performance, exceeding BM by 1.1\% in AP metric.

Qualitative results corresponding to these settings are presented in Fig.~\ref{fig:ablation}. As previously discussed, pseudo labels generated by SGB tend to suffer from under-segmentation (\textit{e.g., multiple chairs in rows (\uppercase\expandafter{\romannumeral2}) and (\uppercase\expandafter{\romannumeral4}) are merged into a single instance}), while those from MGB often lead to over-segmentation (\textit{e.g., individual chairs in rows (\uppercase\expandafter{\romannumeral1}) and (\uppercase\expandafter{\romannumeral3}) are split into multiple instances}). Although the BM strategy attempts to reconcile these two extremes, it fails to effectively balance the trade-off. In contrast, our GAIM strategy produces results that are visually closer to the ground truth and exhibit fewer segmentation artifacts.

Both the quantitative and qualitative comparisons clearly demonstrate that GAIM successfully integrates complementary mask information from two granular levels, achieving a superior balance between under-segmentation and over-segmentation.

\begin{table}[]
\centering
\caption{Ablation results of SSP. $^{*}$ indicates the results of the pseudo labels on the training set, while the remaining represent the prediction results on the validation set. "Sel." denotes "Selection" and "Prop." indicates "Propagation".}
\scalebox{0.92}{
\begin{tabular}{cl|ccccc}
\toprule
                           & {\color[HTML]{000000} Setting}               & {\color[HTML]{000000} mIoU$^{*}$} & {\color[HTML]{000000} mIoU} & {\color[HTML]{000000} AP}   & {\color[HTML]{000000} AP$_{50}$} & {\color[HTML]{000000} AP$_{25}$} \\ \midrule
{\color[HTML]{000000} (a)} & {\color[HTML]{000000} Baseline}              & {\color[HTML]{000000} 60.6}      & {\color[HTML]{000000} 50.9} & {\color[HTML]{000000} 23.6} & {\color[HTML]{000000} 39.7} & {\color[HTML]{000000} 52.5} \\
{\color[HTML]{000000} (b)} & {\color[HTML]{000000} Sel.}             & {\color[HTML]{000000} 63.1}      & {\color[HTML]{000000} 52.5} & {\color[HTML]{000000} 25.0} & {\color[HTML]{000000} 41.3} & {\color[HTML]{000000} 55.6} \\
{\color[HTML]{000000} (c)} & {\color[HTML]{000000} Sel. + Prop.} & {\color[HTML]{000000} \textbf{67.2}}      & {\color[HTML]{000000} \textbf{54.2}} & {\color[HTML]{000000} \textbf{26.5}} & {\color[HTML]{000000} \textbf{43.3}} & {\color[HTML]{000000} \textbf{55.8}} \\ \bottomrule
\end{tabular}}
\vspace{-0.5cm}
\label{tab:ablation_SSP}
\end{table}


{\bf \textit{Effect of SSP.}} Tab.~\ref{tab:ablation_SSP} presents an ablation study evaluating the effectiveness of each component in the Semantic Selection and Propagation (SSP) strategy. In the baseline setting (a), we directly apply the $max$ operation operation on the classification score matrix $\mathbf{S}$ to select the top-$\alpha\%$ points with the highest scores. Compared with (a), setting (b) enhances label quality by selecting the top-$\alpha\%$ points separately for each category. Experimental results demonstrate that setting (b) outperforms setting (a) by 1.6\% and 1.4\% in terms of mIoU and AP metrics on the validation set, respectively.  This indicates that category-wise selection can effectively mitigate the performance degradation arising from the under-representation of minority classes. Building upon setting (b), setting (c) further improves label quality by propagating the selected semantic labels within each superpoint. Results show that setting (c) surpasses setting (b) by 1.7\% and 1.5\% in mIoU and AP, respectively. These gains suggest that leveraging geometric priors via superpoint-based propagation enhances spatial consistency and reduces label fragmentation, thereby promoting more effective model training.

\begin{table}[]
\centering
\caption{Ablation results of Instance Mask Filter (IMF). }
\begin{tabular}{cc|ccc}
\toprule
{\color[HTML]{000000} }& {\color[HTML]{000000} Setting}     & {\color[HTML]{000000} AP}   & {\color[HTML]{000000} AP$_{50}$} & {\color[HTML]{000000} AP$_{25}$} \\ \midrule
{\color[HTML]{000000} (a)} & {\color[HTML]{000000} w/o IMF} & {\color[HTML]{000000} 25.4} & {\color[HTML]{000000} 42.1} & {\color[HTML]{000000} 54.8} \\
{\color[HTML]{000000} (b)} & {\color[HTML]{000000} w IMF}   & {\color[HTML]{000000} \textbf{26.5}} & {\color[HTML]{000000} \textbf{43.3}} & {\color[HTML]{000000} \textbf{55.8}} \\ \bottomrule
\end{tabular}
\vspace{-0.5cm}
\label{tab:ablation_IMF}
\end{table}

{\bf \textit{Effect of Instance Mask Filter.}} To explore the impact of the Instance Mask Filter module, we conduct ablation studies as shown in Tab.~\ref{tab:ablation_IMF}. In setting (a), the proposal features $PF_i$ are directly fed into the scoring network without any filtering. In setting (b), we incorporate the Instance Mask Filter module into the pipeline. The results show consistent improvements across all metrics, indicating that Instance Mask Filter effectively suppresses the noise caused by inconsistencies in pseudo labels. This leads to more reliable instance representations and facilitates better model learning.

\section{Conclusion}
In this paper, we introduce DBGroup, a Dual-Branch Point Grouping framework designed to project both semantic and mask representations from multi-view images into 3D space. Our framework generates multi-granularity instance masks under scene-level label constraints, and we developed two specialized pseudo-label refinement strategies to effectively integrate instance masks and enhance semantic accuracy. Furthermore, we propose an Instance Mask Filtering mechanism to mitigate training noise resulting from label inconsistencies. Extensive experimental evaluations demonstrate that our approach outperforms existing point-level annotation methods in instance segmentation and surpasses scene-level annotation methods in semantic segmentation. We contend that leveraging scene-level labels for weakly supervised segmentation represents a new and promising research direction with significant potential for future development.

\section{Acknowledgements}
This work was supported in part by the National Natural Science Foundation of China under Grants 62371310, 62501403 and W2412099, in part by the Guangdong Basic and Applied Basic Research Foundation under Grant 2023A1515011236, in part by the Shenzhen Science and Technology Program (JCYJ20241202124415021 and KJZD20230923114605011), in part by the Stable Support Project of Shenzhen (Project No.20231122122722001) and in part by the Scientific Foundation for Youth Scholars of Shenzhen University.

\bibliography{aaai2026}


\clearpage
\appendix
\setcounter{table}{0}   
\setcounter{figure}{0}
\setcounter{section}{0}
\setcounter{equation}{0}

\onecolumn{ 
\begin{center}
    \fontsize{20pt}{24pt}\selectfont 
    \textbf{Supplementary Material} 
    \vspace*{0.5cm} 
\end{center}
\addcontentsline{toc}{section}{Supplementary Material} 
}

\begin{multicols}{2}
\section{More Implementations Details}

\subsection{Details of 3D-2D Projection}
Following the approach in \cite{bpnet}, we compute the 2D image coordinates
$(h_{n}, w_{n})$ for any 3D scene point $\mathbf{P}_{n} = (x_n,y_n,z_n)$ by applying the intrinsic camera calibration matrix $\mathcal{I}_{f}$ and the extrinsic camera pose matrix $\mathcal{E}_{f}$ corresponding to the $f$-th frame $\mathbf{I}_{f}$:
\begin{equation}
\frac{1}{d_n} \cdot
\begin{bmatrix}
h_{n} \\[3pt]
w_{n} \\[3pt]
1
\end{bmatrix}
=
\mathcal{I}_{f} \cdot \mathcal{E}_{f} \cdot
\begin{bmatrix}
x_n \\[3pt]
y_n \\[3pt]
z_n \\[3pt]
1
\end{bmatrix},
\end{equation}
where $d_{n}$ denotes the depth projection of the point in the image, and we perform occlusion tests based on the provided depth images $\mathbf{D}_{f}$.


\subsection{Details of Granularity Aware Assignment Strategy}
Algorithm \ref{alg:GAA} presents the details of our Granularity-Aware Assignment algorithm.

\begin{algorithm}[H]
    \caption{Granularity Aware Assignment}
    \label{alg:GAA}
    \renewcommand{\algorithmicrequire}{\textbf{Input:}}
    \renewcommand{\algorithmicensure}{\textbf{Output:}}
    
    \begin{algorithmic}[1]
        \REQUIRE Coarse-grained instance masks $\{\mathbf{M}_{q}\}_{q=1}^{Q}$; \\ \quad \ \  Fine-grained instance masks $\{\mathbf{O}_{w}\}_{w=1}^{W}$; \\ \quad \ \  Overlap threshold $\theta \in [0,1]$;
        \ENSURE Ensemble instance masks $\Gamma =\{\mathbf{E}_{r}\}_{r=1}^{R}$
        \STATE  initialize overlap matrix $\mathbf{A} \in \{0\}^{Q \times W}$
        \STATE  initialize an empty set $\Gamma = \varnothing$
        \FOR{ $q = 1$ \textbf{to} $Q$}
            \FOR{ $w = 1$ \textbf{to} $W$}
                \STATE $A[q][w] = \textbf{sum}(\mathbf{M}_{q} \cap \mathbf{O}_{w})$
            \ENDFOR
        \ENDFOR

        \FOR{ $q = 1$ \textbf{to} $Q$}   
            \STATE $\rho \gets \textbf{max}(A[q,:]) / \textbf{sum}(A[q,:])$  

            \IF {$\rho > \theta$}
                \STATE add $\mathbf{M}_{q}$ to $\Gamma$
            \ELSE
                \FOR{ $w = 1$ \textbf{to} $W$}
                    \STATE add $\mathbf{M}_{q} \cap \mathbf{O}_{w}$ to $\Gamma$
                \ENDFOR
            \ENDIF
        \ENDFOR
        \RETURN $\Gamma$
    \end{algorithmic}
\end{algorithm}

\subsection{Details of Network Architecture}

Our network architecture is illustrated in Fig.4 of the main paper, and we provide detailed descriptions below.

{\bf \textit{Feature Extraction.}} For a point cloud scene $\{\mathbf{P_{n}}\}_{n=1}^{N}$, we first voxelize the points into regular voxels and adopt a sparse 3D U-Net \cite{mink} to extract features and recover
points from voxels obtaining point-wise features $\{\mathbf{F_{n}} \}_{n=1}^{N} \in \mathbb{R}^{N\times C}$.

{\bf \textit{Semantic Branch.}} Point-wise features $\{\mathbf{F_{n}} \}_{n=1}^{N}$ are fed into a three-layer MLP to predict semantic scores $\mathbf{\hat{Y}}^{S} \in \mathbb{R}^{N\times K}$, where $K$ denotes the number of categories in the dataset. The semantic branch is supervised by the cross-entropy loss $L_{sem}$:
\begin{equation}
L_{sem} = -\frac{1}{N}\sum_{i=1}^{N}\text{CE}(\mathbf{\hat{Y}}^{S},\mathbf{Y}^{S}), 
\end{equation}
where $\mathbf{Y}^{S}$ represents the semantic pseudo labels from Semantic Selection and Propagation strategy.

{\bf \textit{Offset Branch.}} In parallel, we use a three-layer MLP to predict the offset for each point $\mathbf{\hat{Y}}^{O} \in \mathbb{R}^{N\times 3}$. Following \cite{pointgroup, softgroup, pbnet}, for points belonging to the same instance, we constrain $\mathbf{\hat{Y}}^{O}$ using an $L_1$ regression loss:
\begin{equation}
L_{off} = \frac{1}{N}\sum_{i=1}^{N}\Vert \mathbf{\hat{Y}}^{O} - \mathbf{Y}^{O} \Vert,
\end{equation}
where $\mathbf{Y}^{O}$ is calculated as the difference between each point and the center point of its corresponding instance, and the instance pseudo label $\mathbf{Y}^{I}$ specifies which point belongs to which instance.

Similar to previous work \cite{pointgroup}, we also adopt a direction loss to encourage each point to move towards the correct direction by constraining the angle between the predicted offset vector and the ground-truth vector:
\begin{equation}
L_{dir} = -\frac{1}{N}\sum_{i=1}^{N}\frac{\mathbf{\hat{Y}}^{O}}{\Vert \mathbf{\hat{Y}}^{O} \Vert_{2}^{row}} \cdot \frac{\mathbf{Y}^{O}}{\Vert \mathbf{Y}^{O} \Vert_{2}^{row}},
\end{equation}
where $\Vert \Vert_{2}^{row}$ denotes the $L_2$ norm computed row-wise.

{\bf \textit{Cluster.}} We apply the point-wise offset $\mathbf{\hat{Y}}^{O}$ to the coordinates of each point, and then use the BFS grouping algorithm described in the main paper to cluster the shifted points with close positions and the same semantics to obtain instance proposals.

{\bf \textit{Scoring.}} Following \cite{pointgroup, pbnet}, we adopt a ScoreNet to evaluate and score all instance proposals. ScoreNet is a lightweight 3D U-net that extracts features from instance proposals and predicts scores $\hat{SC} \in \mathbb{R}^{N_{p}}$ for them, where $N_{p}$ denotes the number of instance proposals. The ScoreNet is supervised by:
\begin{equation}
L_{sc} = -\frac{1}{N_{p}}\sum_{i=1}^{N_{p}}(SC_{i}\log(\hat{SC}_{i}) + (1-SC_{i})\log(1-\hat{SC}_{i})),
\end{equation}
where $SC_{i}$ is calculate by:
\begin{equation}
SC_{i} = \begin{cases}
0 & iou_i < \theta_l \\
1 & iou_i > \theta_h \\
\frac{1}{\theta_h - \theta_l} \cdot (iou_i - \theta_l) & otherwise
\end{cases},
\end{equation}
where $\theta_l$ and $\theta_h$ is set to 0.25 and 0.75, respectively, and $iou_{i}$ is the largest Intersection over Union (IoU) between the $i$-th proposal and all ground truth instances.

\section{Runtime Analysis}
We report the runtime of each component on the ScanNet training set in Tab.~\ref{tab:runtime}, measured using an RTX 4090 GPU and AMD EPYC 9534 64-core processor. Notably, even including the pseudo label generation and optimization time, our total time remains lower than OTOC's annotation time.

\begin{table}[H]
\centering
\caption{Runtime of each component.}
\scalebox{0.9}{
\begin{tabular}{c|ccccc}
\toprule
{\color[HTML]{000000} }    & {\color[HTML]{000000} SGB} & {\color[HTML]{000000} MGB} & {\color[HTML]{000000} GAIN}   & {\color[HTML]{000000} SSP} \\ \midrule
{\color[HTML]{000000} Runtime(s/scene)} & {\color[HTML]{000000} 22.7948}  & {\color[HTML]{000000} 45.0732} & {\color[HTML]{000000} 0.8501} & {\color[HTML]{000000} 0.0784} \\
\bottomrule
\end{tabular}}
\vspace{-0.2cm}
\label{tab:runtime}
\end{table}

\section{More Ablation Study}

{\bf \textit{Ablation on Self-Training Iterations $T$.}} As shown in Tab.\ref{tab:ablation_iterations}, increasing the number of self-training iterations consistently enhances model performance. This trend indicates that additional rounds of self-training progressively refine the pseudo labels, leading to higher-quality supervision signals. We set $T = 3$ as the default configuration, as further increasing the number of iterations yields marginal gains while incurring additional computational overhead.

{\bf \textit{Ablation on Overlap Threshold $\theta$ in GAIM.}} 
The GAIM module partitions coarse-grained instances based on an overlap threshold $\theta$. As shown in Tab.\ref{tab:ablation_theta}, we can find that setting $\theta$ either too low or too high degrades the overall performance. 
Specifically, a lower threshold provides insufficient criteria for effective splitting, resulting in most coarse-grained instances remaining unsegmented. In contrast, an excessively high threshold leads to the generation of numerous fine-grained instances, which may introduce over-segmentation artifacts and compromise instance integrity.

\begin{table}[H]
\centering
\begin{minipage}[t]{0.22\textwidth} 
\centering
\captionsetup{width=\linewidth} 
\caption{Ablation results of self-training iterations.}
\setlength{\tabcolsep}{1.4mm}{

\begin{tabular}{c|ccc}
\toprule
{\color[HTML]{000000} $T$} & {\color[HTML]{000000} AP}            & {\color[HTML]{000000} AP$_{50}$}          & {\color[HTML]{000000} AP$_{25}$}          \\ \midrule
{\color[HTML]{000000} 1}                  & {\color[HTML]{000000} 26.5}          & {\color[HTML]{000000} 43.3}          & {\color[HTML]{000000} 55.8}          \\
{\color[HTML]{000000} 2}                  & {\color[HTML]{000000} 28.6}          & {\color[HTML]{000000} 44.9}          & {\color[HTML]{000000} 58.4}          \\
{\color[HTML]{000000} 3}                  & {\color[HTML]{000000} 28.6} & {\color[HTML]{000000} \textbf{46.2}} & {\color[HTML]{000000} \textbf{59.6}} \\
{\color[HTML]{000000} 4}                  & {\color[HTML]{000000} 28.6}          & {\color[HTML]{000000} 45.9}          & {\color[HTML]{000000} 58.8}          \\
{\color[HTML]{000000} 5}                  & {\color[HTML]{000000} \textbf{28.8}}          & {\color[HTML]{000000} 45.5}          & {\color[HTML]{000000} 58.4}          \\ \bottomrule
\end{tabular}

}

\label{tab:ablation_iterations}

\end{minipage}
\hfill 
\begin{minipage}[t]{0.22\textwidth} 
\centering
\captionsetup{width=\linewidth} 
\caption{Ablation results of overlap threshold in GAIM.}
\setlength{\tabcolsep}{1.4mm}{

\begin{tabular}{c|ccc}
\toprule
{\color[HTML]{000000} $\theta$}                         & {\color[HTML]{000000} AP}            & {\color[HTML]{000000} AP$_{50}$}          & {\color[HTML]{000000} AP$_{25}$}          \\ \midrule
{\color[HTML]{000000} 0.2} & {\color[HTML]{000000} 17.1}          & {\color[HTML]{000000} 32.4}          & {\color[HTML]{000000} 57.0}          \\
{\color[HTML]{000000} 0.3} & {\color[HTML]{000000} 17.3}          & {\color[HTML]{000000} 32.5}          & {\color[HTML]{000000} 57.7}          \\
{\color[HTML]{000000} 0.4} & {\color[HTML]{000000} \textbf{17.4}} & {\color[HTML]{000000} \textbf{32.7}} & {\color[HTML]{000000} 58.7} \\
{\color[HTML]{000000} 0.5} & {\color[HTML]{000000} 17.2}          & {\color[HTML]{000000} 32.5}          & {\color[HTML]{000000} 59.0}          \\
{\color[HTML]{000000} 0.6} & {\color[HTML]{000000} 16.7}          & {\color[HTML]{000000} 32.1}          & {\color[HTML]{000000} \textbf{59.1}}          \\ \bottomrule
\end{tabular}

}

\vspace{-0.2cm}
\label{tab:ablation_theta}
\end{minipage}
\end{table}

{\bf \textit{Ablation on grouping strategy in SGB.}} We compared the performance of BFS grouping \cite{pointgroup} and DBSCAN \cite{dbscan}. For fair comparison, we perform DBSCAN on a per-class basis. As demonstrated in Tab.\ref{tab:grouping}, BFS grouping achieves superior performance over DBSCAN in terms of both performance and computational efficiency.

\begin{table}[H]
\centering
\caption{Ablation results of different grouping strategy.}
\scalebox{0.82}{
\begin{tabular}{c|ccccc}
\toprule
{\color[HTML]{000000} Method}    & {\color[HTML]{000000} AP} & {\color[HTML]{000000} $\text{AP}_{50}$} & {\color[HTML]{000000} $\text{AP}_{25}$}   & {\color[HTML]{000000} Runtime (s/scene)} \\ \midrule
{\color[HTML]{000000} DBSCAN (Class-wise)} & {\color[HTML]{000000} 17.1}  & {\color[HTML]{000000} 31.5} & {\color[HTML]{000000} 54.9} & {\color[HTML]{000000} 0.4031} \\
{\color[HTML]{000000} BFS (Ours)} & {\color[HTML]{000000} \textbf{17.4}}  & {\color[HTML]{000000} \textbf{32.7}} & {\color[HTML]{000000} \textbf{58.7}} & {\color[HTML]{000000} \textbf{0.1459}} \\
\bottomrule
\end{tabular}}
\label{tab:grouping}
\end{table}

{\bf \textit{Ablation on top-$\alpha\%$ threshold in SSP.}} We further investigate the impact of the top-$\alpha\%$ threshold in SSP, as shown in Tab.\ref{tab:ablation_alpha}. 
Regarding the pseudo labels on the training dataset, we observe that as the threshold $\alpha$ increases, the mIoU of the pseudo labels progressively decreases. This indicates that a higher confidence level in SSP yields pseudo labels of better quality. For the performance on the validation dataset, the mIoU initially improves with an increasing supervision rate, but starts to decline once $\alpha$ exceeds 30\%. When $\alpha$ is too small, only a limited number of pseudo labels are retained, which leads to insufficient supervision and suboptimal segmentation performance. Conversely, when $\alpha$ is too large, more noisy pseudo labels are retained, misleading the model and degrading its performance. To strike a balance between supervision quantity and label quality, we choose 30\% as the optimal top-$\alpha\%$ threshold for SSP.

\vspace{-0.2cm}

\begin{table}[H]
\centering
\caption{Ablation results of top-$\alpha\%$ threshold in SSP. $^{*}$ indicates the results of the pseudo labels on the training set.}
\begin{tabular}{c|ccccc}
\toprule
{\color[HTML]{000000} $\alpha (\%)$}    & {\color[HTML]{000000} mIoU*} & {\color[HTML]{000000} mIoU} & {\color[HTML]{000000} AP}   & {\color[HTML]{000000} AP$_{50}$} & {\color[HTML]{000000} AP$_{25}$} \\ \midrule
{\color[HTML]{000000} 10} & {\color[HTML]{000000} \textbf{68.8}}  & {\color[HTML]{000000} 47.6} & {\color[HTML]{000000} 21.4} & {\color[HTML]{000000} 36.5} & {\color[HTML]{000000} 52.8} \\
{\color[HTML]{000000} 20} & {\color[HTML]{000000} 68.0}  & {\color[HTML]{000000} 52.0} & {\color[HTML]{000000} 24.1} & {\color[HTML]{000000} 40.4} & {\color[HTML]{000000} 55.3} \\
{\color[HTML]{000000} 30} & {\color[HTML]{000000} 67.4}  & {\color[HTML]{000000} \textbf{54.2}} & {\color[HTML]{000000} \textbf{26.5}} & {\color[HTML]{000000} \textbf{43.3}} & {\color[HTML]{000000} \textbf{55.8}} \\
{\color[HTML]{000000} 40} & {\color[HTML]{000000} 66.9}  & {\color[HTML]{000000} 53.8} & {\color[HTML]{000000} 25.8} & {\color[HTML]{000000} 41.5} & {\color[HTML]{000000} 55.1} \\
{\color[HTML]{000000} 50} & {\color[HTML]{000000} 65.6}  & {\color[HTML]{000000} 53.1} & {\color[HTML]{000000} 25.1} & {\color[HTML]{000000} 41.4} & {\color[HTML]{000000} 55.2} \\ \bottomrule
\end{tabular}
\vspace{-0.2cm}
\label{tab:ablation_alpha}
\end{table}


\section{Limitations}

The quality of our pseudo labels is inherently constrained by the performance of both the semantic and mask branches. If either branch fails to accurately identify the target, the resulting pseudo labels will inevitably suffer in quality. Moreover, our method relies on superpoints generated through the over-segmentation of 3D point clouds, which introduces additional computational overhead and increases resource consumption.

\vspace{-0.2cm}


\end{multicols}

\end{document}